\documentclass[lettersize,journal]{IEEEtran}
\usepackage{amsmath,amsfonts}
\usepackage{algorithm}
\usepackage{array}
\usepackage[caption=false,font=normalsize,labelfont=sf,textfont=sf]{subfig}
\usepackage{textcomp}
\usepackage{stfloats}
\usepackage{url}
\usepackage{verbatim}
\usepackage{graphicx}
\usepackage{cite}
\usepackage{graphicx}
\usepackage{amsmath}
\usepackage{amssymb}
\usepackage{booktabs}
\usepackage{mathtools}
\usepackage{xcolor}
\usepackage{xspace}
\usepackage[pagebackref,breaklinks,colorlinks]{hyperref}
\usepackage{booktabs}

\makeatletter
\DeclareRobustCommand\onedot{\futurelet\@let@token\@onedot}
\def\@onedot{\ifx\@let@token.\else.\null\fi\xspace}

\def\eg{\emph{e.g}\onedot} 
\def\ie{\emph{i.e}\onedot}

\def\etal{\emph{et al}\onedot}
\makeatother

\usepackage[capitalize]{cleveref}
\crefname{section}{Sec.}{Secs.}
\Crefname{section}{Section}{Sections}
\Crefname{table}{Table}{Tables}
\crefname{table}{Tab.}{Tabs.}

\usepackage{algorithm}
\usepackage{algpseudocode}
\usepackage{amssymb}
\begin{document}

\title{RW-Net: Enhancing Few-Shot Point Cloud Classification with a Wavelet Transform Projection-based Network}

\author{Haosheng Zhang$^\dagger$, Hao Huang$^\dagger$
\thanks{$\dagger$ indicates equal contribution.\\
Haosheng Zhang is with NYU Tandon School of Engineering, New York University, USA; Hao Huang is with NYUAD Center for Artificial Intelligence and Robotics, New York University Abu Dhabi, UAE. \\
Email:\{hz3145,hh1811\}@nyu.edu.}
}
        % <-this % stops a space

\maketitle

\begin{abstract}
In the domain of 3D object classification, a fundamental challenge lies in addressing the scarcity of labeled data, which limits the applicability of traditional data-intensive learning paradigms. This challenge is particularly pronounced in few-shot learning scenarios, where the objective is to achieve robust generalization from minimal annotated samples. To overcome these limitations, it is crucial to identify and leverage the most \textit{salient} and \textit{discriminative} features of 3D objects, thereby enhancing learning efficiency and reducing dependency on large-scale labeled datasets. In this work, we introduce RW-Net, a novel framework designed to address the aforementioned challenges by integrating \underline{R}ate-Distortion Explanation (RDE) and \underline{w}avelet transform into a state-of-the-art projection-based 3D object classification architecture. The proposed method capitalizes on RDE to extract critical features by identifying and preserving the most \textit{informative} data components while simultaneously reducing redundancy. This process ensures the retention of essential information for effective decision-making, thereby optimizing the model's ability to learn from limited data. Complementing RDE, the incorporation of the wavelet transform further enhances the framework's capability to generalize in low-data regimes. By emphasizing low-frequency components of the input data, the wavelet transform captures fundamental \textit{geometric} and \textit{structural} attributes of 3D objects. These attributes are instrumental in mitigating overfitting and improving the robustness of the learned representations across diverse tasks and domains. To validate the effectiveness of our RW-Net, we conduct extensive experiments on three benchmark datasets: ModelNet40, ModelNet40-C, and ScanObjectNN for few-shot 3D object classification. The results demonstrate that our approach not only achieves state-of-the-art performance but also exhibits superior generalization and robustness in few-shot learning scenarios. Furthermore, we observe consistent improvements in classification accuracy, even under challenging conditions such as data corruption and cross-domain variations. We will release our code upon acceptance, contributing a tool to the research community for advancing few-shot 3D object classification.
\end{abstract}
      
%%%%%%%%%%%%%%%%%%%%%%%%%%%%%%%%%%%%%%%%%%%%%%%%%%%%%%%%%%%%%%%%%%%%%%%%%%%%%%%%%%%%%%%%%%%%%%%%%%%%%%%%
\section{Introduction}
\IEEEPARstart{P}{oint} clouds offer rich geometric data, enabling precise modeling of three-dimensional structures and environments, which is crucial for vision tasks, including object detection~\cite{zhou2018voxelnet,lu2023open}, recognition~\cite{komorowski2021minkloc3d,dong2023inor}, and scene understanding~\cite{jaritz2019multi,chen2023clip2scene}. The importance of point cloud learning has therefore grown significantly due to its wide applications in fields such as autonomous driving~\cite{yang2024visual,yuan2024ad} and robotics~\cite{ni2020pointnet++,christen2023learning}. Few-shot learning in point cloud processing is particularly important, as it addresses the challenge of limited labeled data for training, which is often expensive and labor-intensive to acquire. Therefore, few-shot learning allows a model to generalize from a small number of training examples, making it possible to perform well on new and unseen data. However, few-shot learning in the context of 3D point clouds faces several challenges~\cite{pourpanah2022review,ye2023closer,xiao2024survey}, including but not limited to, the high dimensionality and irregularity of point cloud data, the need for effective data augmentation, and the difficulty in maintaining the robustness and accuracy of models when trained on sparse data. Addressing these challenges is essential for enabling more efficient and scalable 3D point cloud few-shot learning.

% Existing methods of few-shot learning for 2D images
% Existing methods of few-shot learning for 3D point clouds

Note that few-shot learning for 2D images has made significant strides with various techniques proposed to leverage limited labeled data efficiently. Predominant approaches include meta-learning~\cite{hospedales2021meta,huisman2021survey,he2023few}, which trains models to quickly adapt to new tasks with only a few examples by simulating the few-shot learning scenario during training. Gradient-based methods~\cite{finn2017model,lee2019meta,lee2022contextual} and metric-based methods~\cite{snell2017prototypical,mai2019attentive,chen2020variational} are notable examples. The former involves training models in a way that allows them to quickly adapt to new tasks with only a few gradient updates by optimizing initial network parameters that are easily fine-tuned, while the latter focuses on learning a distance metric to compare new samples with a small set of labeled examples. In contrast, few-shot learning for 3D point clouds is still emerging, but innovative adaptations of the above methods have been seen. For instance,~\cite{puri2020few,huang20213d} generate task-specific initial parameters of a network, allowing rapid adaptation to new tasks with minimal labeled data, while~\cite{ye2022makes,anvekar2023gpr} learn a metric to measure the similarity between pairs of images, facilitating classification with minimal data. Additionally, a recent work ViewNet~\cite{anvekar2023gpr} identifies the limitations of point-based backbones, \eg, PointNet~\cite{qi2017pointnet} and DGCNN~\cite{wang2019dynamic}, in few-shot learning for 3D point cloud classification, particularly their sensitivity to occlusions and missing points. Therefore, ViewNet employs a projection-based backbone that transforms 3D point clouds into multiple 2D depth images from different views and introduces a view pooling mechanism to combine the features from these projections, ensuring robustness against missing points. A more recent work SimpliMix~\cite{yang2024SimpliMix} alleviates the overfitting issue in few-shot point cloud classification due to the small size and lack of diversity in 3D point cloud datasets, which contrasts with the larger and more diverse 2D image datasets, by linearly interpolating pairs of hidden representations of point clouds and then mixing them within each training episode, forcing the model to learn more robust features that generalize better to novel classes.

We notice that in few-shot learning, the scarcity of training data necessitates models to efficiently capture the most \textit{salient} features of the input to generalize effectively to novel classes. To align with the need to extract the most informative features from limited data, we propose RW-Net which systematically adapts and integrates \textit{Rate-Distortion Explanation} (RDE)~\cite{macdonald2019rate,kolek2022cartoon} and \textit{wavelet transform}~\cite{zhao2022wavelet} with ViewNet~\cite{chen2023viewnet}, a state-of-the-art projection-based 3D point cloud few-shot learning model. The RDE emphasizes learning a sparse and efficient representation of the input that preserves the essential features while discarding unnecessary details, which enables the model to learn efficiently with limited training data and generalize well to new tasks. Specifically, we produce piece-wise smooth projection images by enforcing sparsity in the wavelet domain, only capturing the most relevant parts of an image that lead to a model's decision while minimizing distortion in the model output. Furthermore, as indicated in~\cite{wang2020high}, humans typically use low-frequency components of an image, which mainly contain texture and semantic information, for object recognition, whereas conventional neural networks process a blend of low- and high-frequency components. Since high-frequency components consist of both pertinent data details and noise, a mixture of high-frequency components without filtering out noise may potentially cause overfitting during training. Therefore, we incorporate wavelet transform into a 2D backbone network to let it focus on the low-frequency components of the input, making the learning process more robust and effective, thus enhancing its ability to generalize to new classes in few-shot scenarios. To evaluate the effectiveness of our RW-Net, we conducted experiments on three widely used datasets with cross-validation: ModelNet40~\cite{wu20153d}, ModelNet40-C~\cite{sun2022benchmarking}, and ScanObjectNN~\cite{uy2019revisiting}. The experimental results show that our method consistently exceeds the state-of-the-art performance in few-shot point cloud classification. The main contributions of this work are summarized as follows:
\begin{itemize}
    \item We adapt the Rate-Distortion Explanation (RDE) to enhance learning efficiency and generalization in 3D few-shot classification by learning a sparse and efficient representation of input data, preserving essential features while minimizing unnecessary details.
    \item We incorporate the wavelet transform into a 2D backbone network to focus on low-frequency components, capturing relevant parts of an image and minimizing noise, thereby making the learning process more robust and reducing the risk of overfitting.
    \item We evaluate the effectiveness of our proposed method through experiments on three public datasets, \ie, ModelNet40, ModelNet40-C, and ScanObjectNN, on which it consistently achieves state-of-the-art performance in few-shot point cloud classification.
\end{itemize}
%

%%%%%%%%%%%%%%%%%%%%%%%%%%%%%%%%%%%%%%%%%%%%%%%%%%%%%%%%%%%%%%%%%%%%%%%%%%%%%%%%%%%%%%%%%%%%%%%%%%%%%%%%
\section{Related Work}
\noindent 
\textbf{Point cloud classification.}
Point cloud classification is crucial in areas such as autonomous driving~\cite{yang2024visual,yuan2024ad}, robotics~\cite{ni2020pointnet++,christen2023learning}, and 3D modeling~\cite{dong2023inor,lu2023open,chen2023clip2scene}, significantly advanced by deep learning. PointNet~\cite{qi2017pointnet} and PointNet++~\cite{ni2020pointnet++} pioneered this field by processing raw point clouds and introducing hierarchical feature learning. Multiview approaches have also been influential, with MVCNN~\cite{su2015multi} processing multiple 2D views for feature aggregation, and MVP~\cite{jaritz2019multi} combining PointNet with multiview representations. Recent innovations include ViewNet~\cite{chen2023viewnet}, which synthesizes features across views, and PointFormer~\cite{chen2022pointformer} that applies self-attention for accuracy improvements. Other notable developments include 3DCTN~\cite{lu20223dctn} which merges 3D convolutions with transformers, CLIP2Point~\cite{huang2023clip2point} which adapts the CLIP model~\cite{radford2021learning} to point clouds, ERINet~\cite{weng2022erinet} which addresses rotation variance in point cloud, and the geometry sharing network~\cite{xu2020geometry} which captures intrinsic structures. We build our work on ViewNet due to its balance between classification accuracy and computational cost.

\noindent \textbf{3D few-shot learning.}
Few-shot learning, training models with limited data, has extended to 3D data recently, building on foundational 2D image methods like Matching Networks~\cite{vinyals2016matching} and Prototypical Networks~\cite{snell2017prototypical}. These methods set the stage for adapting few-shot learning to 3D contexts, with advancements by Wang \etal~\cite{wang2022relation} which extends Relation Networks~\cite{sung2018learning} for 3D point clouds. Recent innovations in 3D few-shot learning include disentangled Prototypical Networks~\cite{prabhudesai2021disentangling}, UVStyle-Net~\cite{meltzer2021uvstyle} using unsupervised methods for style similarity, and FILP-3D~\cite{xu2023filp} leveraging pre-trained vision-language models. Further developments include AgileGAN3D~\cite{song2024agilegan3d} applying augmented transfer learning, few-shot class-incremental learning by Cheraghian \etal~\cite{cheraghian2021synthesized}, a prototypical variational autoencoder by Hu \etal~\cite{tang2023prototypical}, CaesarNeRF~\cite{zhu2024caesarnerf} using semantic representations for neural rendering, and MetaFSCIL~\cite{chi2022metafscil} employing meta-learning for scalability. In addition, Zhu \etal~\cite{zhu2023less} proposes training-free networks for semantic segmentation, and Han \etal~\cite{han2024multi} explores multi-animal pose estimation with a few-shot learning framework. In this work, we focus on 3D few-shot classification, which is a fundamental task across multiple real-world vision-related applications.

\noindent \textbf{Wavelet in neural networks.}
The integration of wavelet transforms with neural networks has enhanced feature extraction and learning efficiencies across various fields. A significant work, WaveletCNNs~\cite{rippel2015spectral}, introduces wavelet convolutional neural networks, which incorporate wavelet transforms within the convolutional layers. This approach allowed the model to inherently perform multi-resolution analysis, significantly improving performance on image processing tasks. Another notable work is by Fujieda \etal~\cite{fujieda2018wavelet} that develops a wavelet convolutional neural network that employs wavelet transform as a replacement for the traditional convolution operation in neural networks. The following works~\cite{liu2019multi, jia2022representing,shi2021deep,huang2021adaptive,nguyen2023fast} further model 2D image, 3D point cloud, and even high-dimensional data dynamics using multi-level and non-redundant wavelet transform. In data-intensive environments, Eduru \etal~\cite{eduru2024parallel} implement parallel and streaming wavelet networks on Apache Spark for efficient big data processing, and Lan \etal~\cite{lan2022wavelet} improve fault diagnosis in mechanical systems with a multi-label wavelet network. In cybersecurity, Li \etal~\cite{li2021capability} and Fan \etal~\cite{fan2020soft} utilize wavelet networks for DDoS detection and data denoising, enhancing security measures. In this work, we instead fuse wavelet transform into a Transformer~\cite{vaswani2017attention} to learn discriminative 3D shape features.

%%%%%%%%%%%%%%%%%%%%%%%%%%%%%%%%%%%%%%%%%%%%%%%%%%%%%%%%%%%%%%%%%%%%%%%%%%%%%%%%%%%%%%%%%%%%%%%%%%%%%%%%
\section{Preliminary}
\label{sec:pre}
\noindent \textbf{Rate-Distortion Exploration (RDE).} The components of an input signal $x$ can be divided into a relevant subset $S$ and the remaining non-relevant subset $\bar{S}$. Thus, we can denote $x = x_S + x_{\bar{S}}$. For a given classification model $f$, the relevant subset $S$ of components is determined by its ability to approximate the output of $f(x)$ even when other components are randomized. This is quantified using a probability distribution $\mathcal{V}$ in the input space, where the \textit{obfuscation} of $x$ relative to $S$ and $\mathcal{V}$ involves fixing $S$ while randomizing the remaining complement, \ie, $y = x_S + n$ where $n \sim \mathcal{V}$ is a random vector. The change in the output of $f$ due to this obfuscation is measured by a distance metric, leading to a \textit{rate-distortion} trade-off. This trade-off is a foundational concept in information theory~\cite{cover1999elements}, particularly in the context of lossy data compression, suggesting that the subset $S$ represents a compact, yet effective, representation of the input signal $x$ relative to the decision-making process of the classification model $f$. RDE~\cite{macdonald2019rate} seeks the smallest subset $S$ that keeps the distortion, \ie, the deviation from the original output of the classification model, below a certain threshold $\epsilon$. Specifically, the rate-distortion function is defined as:
\begin{equation}
    R(\epsilon) = \min\{\vert S\rvert: S \subseteq x \wedge D(S) \leq \epsilon\}\enspace,
\end{equation}
where the distance metric $D(S)$ has the definition of:
\begin{equation}
    D(S) = \mathbb{E}_{n \sim \mathcal{V}}\Big[\frac{1}{2}\big(f(x) - f(y)\big)^2\Big] \enspace \text{and}\enspace y = x_S + n\enspace.
\end{equation}
Suppose $x \in \mathbb{R}^d$ and the relevant subset $S$ to be represented by a continuous mask $s \in [0, 1]^d$, RDE \cite{macdonald2019rate} aims to solve the following optimization problem:
\begin{equation}
    \min_{s\in [0, 1]^n}\mathbb{E}_{n \sim \mathcal{V}}\Big[\frac{1}{2}\Big(f(x) - f\big(x\odot s + n \odot (1 - s)\big)\Big)^2\Big] + \lambda \lVert s \rVert_1\enspace,
    \label{eq:s_opt}
\end{equation}
where $\odot$ denotes Hadamard product and $\lambda > 0$ is a hyperparameter controlling the sparsity level of the mask. The following work CartoonX~\cite{kolek2022cartoon} extends RDE from spatial domain (\eg, pixel domain for 2D images) of an input to its spectral domain. Let $\mathcal{F}$ represent discrete wavelet transform and $\mathcal{F}^{-1}$ be its inverse, the obfuscation of $x$ is defined in wavelet domain: 
\begin{equation}
    y \coloneq \mathcal{F}^{-1}\big(\mathcal{F}(x) \odot s + n \odot (1 - s)\big)\enspace.
\end{equation}
A key advantage of generating obfuscation in the wavelet domain is that it allows for the extraction of piece-wise smooth parts of an image, which are more sparse in the wavelet domain, leading to more effective and relevant explanations of classification model decisions.

\begin{figure*}[!htb]
    \centering
    \includegraphics[width=.99\textwidth]{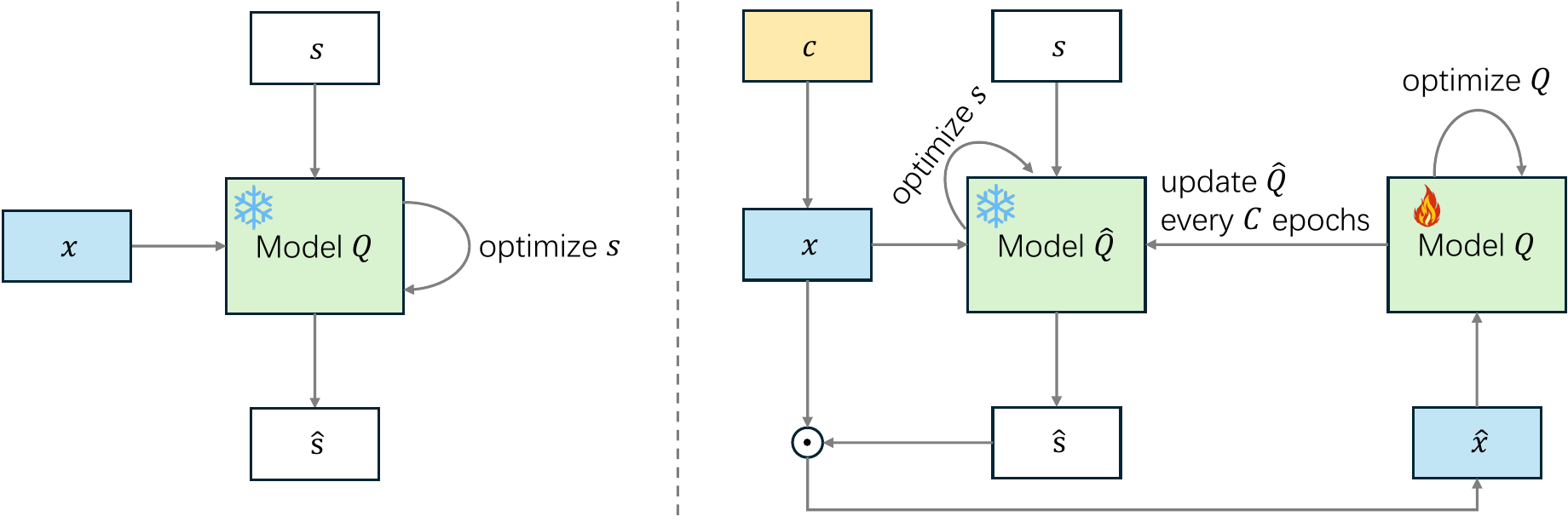}
    \caption{Left: The workflow of CartoonX~\cite{kolek2022cartoon} where $x$ is an input image and $s$ is a continuous mask initialized to be all ones. Right: Our method jointly optimizes image mask and classification model parameters. The symbol $c$ represents a given point cloud, $x$ denotes (one of) the projected images from $c$, and $\odot$ denotes Hadamard product.}
    \label{fig:compare}
\end{figure*}

\noindent \textbf{Discrete wavelet transform.} 
Wavelet transform \cite{daubechies1992ten} provides a mathematical framework for decomposing signals. By employing a set of orthogonal mother wavelets that decay rapidly, wavelet transform permits the dissection of signals into varying frequency and time resolutions by using different scale and translation parameters. To accommodate the discrete nature of 2D images, the Discrete Wavelet Transform (DWT) has been developed, facilitating the decomposition of images at different frequency bands. The mother wavelet of DWT is mathematically represented as:
\begin{equation}
    \psi_{j,k}(t) = \frac{1}{\sqrt{2^j}}\psi(\frac{t-k2^j}{2^j})\enspace,
    \label{eq:dwt}
\end{equation}
where $k$ and $j$ are the translation and scale parameters, both of which are integers. For practical applications in image processing, the two-dimensional DWT extends this concept by decomposing an image into a combination of low-frequency and high-frequency components along horizontal, vertical, and diagonal directions. This can be achieved by applying one-dimensional DWT to the original image's rows, followed by applying it again to the transformed image's columns. Following~\cite{li2020wavelet}, by defining two cyclic matrices, \ie, $\mathbf{L}$ as low-pass filter and $\mathbf{H}$ as high-pass filter:
\begin{equation}
\mathbf{L} = 
\begin{pmatrix}
\cdots & \cdots  & \cdots &  &  &  &  \\
 & \cdots & l_0 & l_1 & \cdots  &  &  \\
 &  &  & \cdots & l_0 & l_1 & \cdots  \\
 &  &  &  &  & \cdots  & \cdots 
\end{pmatrix} \in \mathbb{R}^{\lfloor N/ 2 \rfloor \times N}\enspace,
\label{eq:low_pass}
\end{equation}
\begin{equation}
\mathbf{H} = 
\begin{pmatrix}
\cdots & \cdots  & \cdots &  &  &  &  \\
 & \cdots & h_0 & h_1 & \cdots  &  &  \\
 &  &   & \cdots & h_0 & h_1 & \cdots  \\
 &  &  &  &  & \cdots  & \cdots 
\end{pmatrix} \in \mathbb{R}^{\lfloor N/ 2 \rfloor \times N}\enspace,
\label{eq:high_pass}
\end{equation}
where $N$ is the spatial dimension of the input image. By applying the filters to an input image $\mathbf{I}$, we can decompose the image into four sub-bands~\cite{li2020wavelet}:
\begin{equation}
\begin{aligned}
    \mathbf{I}_{ll} = \mathbf{L} \mathbf{I} \mathbf{L}^\top\enspace, \quad& \mathbf{I}_{lh} = \mathbf{H} \mathbf{I} \mathbf{L}^\top\enspace, \\
    \mathbf{I}_{hl} = \mathbf{L} \mathbf{I} \mathbf{H}^\top\enspace, \quad& \mathbf{I}_{hh} = \mathbf{H} \mathbf{I} \mathbf{H}^\top\enspace,
\end{aligned}
\label{eq:decompose}
\end{equation}
where $\mathbf{I}_{ll}$ denotes the low-frequency component and $\mathbf{I}_{lh}, \mathbf{I}_{hl}, \mathbf{I}_{hh}$ represent three high-frequency components. In this work, we adopt Haar wavelet, \ie, the low-pass filter is $\{l_k\}_{k \in \{0, 1\}} = \left\{\frac{1}{\sqrt{2}}, \frac{1}{\sqrt{2}}\right\}$, and the high-pass filter is $\{h_k\}_{k \in \{0, 1\}} = \left\{\frac{1}{\sqrt{2}}, -\frac{1}{\sqrt{2}}\right\}$, to convert an input image from spatial domain to spectral domain.

%%%%%%%%%%%%%%%%%%%%%%%%%%%%%%%%%%%%%%%%%%%%%%%%%%%%%%%%%%%%%%%%%%%%%%%%%%%%%%%%%%%%%%%%%%%%%%%%%%%%%%%%
\section{Method}
In this section, we first provide the motivation of our method in Section~\ref{subsec:Motivation} and describe CartoonX~\cite{kolek2022cartoon} briefly in Section~\ref{subsec:CartoonX}, and then give an overview of our method in Section~\ref{subsec:overview}, followed by a detailed description of our model architecture in Section~\ref{subsec:arch}, and lastly depict the structure of the wavelet attention block in our model in Section~\ref{subsec:wavelet}.

%=======================================================================================================
\subsection{Motivation}
\label{subsec:Motivation}
In this work, we integrate Rate-Distortion Explanation (RDE) and wavelet transform into a projection-based model to address the challenge of limited labeled data. RDE distills the most important features by identifying essential components of the data, and wavelet transform emphasizes low-frequency components that capture fundamental shape attributes. The motivation is to increase learning efficiency and robustness in few-shot learning scenarios by reducing redundancy, preserving important information, and ensuring better generalization on 3D object datasets where labeled data is scarce. The wavelet transform offers a mathematical framework for signal decomposition and has demonstrated significant capabilities in feature extraction for image processing. Conversely, RDE as an explanation algorithm provides insights into the decision-making process of image classification through image masking, facilitating the distillation of information. However, the application of explanation algorithms in the learning process is still unexplored. Therefore, we aim to integrate these techniques to develop a new learnable model for few-shot 3D shape classification.

%=======================================================================================================
\subsection{CartoonX}
\label{subsec:CartoonX}
CartoonX\cite{kolek2022cartoon} is a model-agnostic explanation method for image classifier, based on the RDE~\cite{macdonald2019rate} framework applied in the wavelet domain. An input image \( x \) is transformed into wavelet coefficients \( h \) using the Discrete Wavelet Transform (DWT). The goal is to find a sparse mask \( s \) that minimizes the distortion in the model's output. The optimization problem is formulated as:
\begin{equation}
\min_{s \in \{0,1\}^n : \|s\|_0 \leq \ell} \mathbb{E}_{v \sim V} \left[D \left( \Phi(x), \Phi(\mathcal{T}(h \odot s + (1 - s) \odot v)) \right) \right] \enspace,
\label{eq:opt_bin}
\end{equation}
where \( \Phi \) is a \textit{pre-trained} classification model, \( v \) is a perturbation vector, \( \mathcal{T} \) is the inverse DWT, \( \odot \) denotes element-wise multiplication, and \(D(\cdot, \cdot) \) measures distortion. To make the optimization feasible, the binary mask $s$ is relaxed to continuous values:
\begin{equation}
\min_{s \in [0,1]^n} \mathbb{E}_{v \sim V} \left[D \left( \Phi(x), \Phi(\mathcal{T}(h \odot s + (1 - s) \odot v)) \right) \right] + \lambda \|s\|_1\enspace,
\label{eq:opt_cont}
\end{equation}
where $\|\cdot\|$ denotes $\ell_1$ norm. The workflow of CartoonX is shown in the left panel of Figure~\ref{fig:compare} in which the model $Q$ is frozen during the optimization of $s$. Note that the original CartoonX is only applied to explain the classification decision process within a \textit{pre-trained} model, instead of training a classification model from scratch.

%=======================================================================================================
\subsection{Overview}
\label{subsec:overview}
We propose RW-Net, as depicted on the right panel of Figure~\ref{fig:compare}, which is developed based on CartoonX~\cite{kolek2022cartoon}. As mentioned above, in CartoonX, with a \textit{pre-trained} classification model $Q$ and a group of input images $x \in \mathbb{R}^{3 \times h \times w}$, it optimizes the parameter of a mask $s \in [0, 1]^{h \times w}$ according to Eq.~\ref{eq:opt_cont}, where $h$ and $w$ denote the spatial dimensions of the image. Note that during optimization, the parameters of $Q$ are frozen. The resultant output $\hat{s}$ is utilized to obscure unnecessary parts of the input $x$, enhancing the efficiency of our training process. That is, the mask $\hat{s}$ is applied to $x$ using Hadamard product, resulting in a masked input $\hat{x}$, which is used for classification.

We reasonably argue that for few-shot learning, as we only have access to limited data, it is essential for us to only learn the saliency or necessary image features to classify these images and discard the unnecessary parts. Therefore, we propose our model based on an adaptation of CartoonX. Given a point cloud $c$, we first project it into six images along each axis, from top, bottom, front, back, left, and right~\cite{chen2023viewnet}, without any bells-and-whistles operations. For ease of presentation, we denote one of the projected images as $x$, and the following operations can be applied to all the other projected images in a similar way. Then, we initialize two classification models $\hat{Q}$ and $Q$ with an identical structure. Distinct from CartoonX, we pre-train model $\hat{Q}$ for $M^\prime$ epochs using unmasked image $x$. Once $\hat{Q}$ has been well trained, we freeze it and initialize a mask $s$. The mask $s$ is optimized using $\hat{Q}$ according to Eq.~\ref{eq:opt_cont}. The optimized mask $\hat{s}$ is then multiplied with the original input image $x$ in an element-wise manner to get $\hat{x}$ in which the unnecessary parts have been filtered out. The model $Q$ is optimized using $\hat{x}$ for classification. To continuously refine our model, inspired by~\cite{mnih2015human}, we update the trained model $\hat{Q}$ every $C$ epochs by copying the parameters of $Q$ into $\hat{Q}$. This iterative process allows us to further distill the learned information and enhance the model’s performance. A sketchy pseudocode of our algorithm is provided in Algorithm~\ref{alg:sketchy}, and the full algorithm can be found in Algorithm~\ref{alg:full}.
% \textit{Supplementary Material}.

%
\begin{algorithm}
\caption{Sketchy pseudocode for RW-Net training.}
\label{alg:sketchy}
% \small
\label{alg:few_algo}
\begin{algorithmic}[1]
\State Initialize pre-training epoch $M^\prime$, training epoch $M$, parameter copying step $C$.
\State Initialize classification model \(Q\) and target model \(\hat{Q}\) with the same structure.
\State Initialize image mask $s$ to all ones.
\For{$i\gets 1$ to $M^\prime$ epochs}
\color{gray}
\Procedure{Pre-train model \(\hat{Q}\)}{}:
\State Project point cloud  \(c\) into image \(x\)
\State Forward \(x\) into  model \(\hat{Q}\)
\EndProcedure
\color{black}
\EndFor
% \red{\Comment{The relation between $Q$ and $\hat{Q}$ is unclear.}}
\For{$i\gets 1$ to $M$ epochs} 
% \For{$j\gets 1$ to $N$ batches} 
\If{training model}
\color{gray}
\Procedure{train model \(Q\)}{}:
\For{$j\gets 1$ to $N$ batches} 
\State Project point cloud \(c\) into images \(x\)
\State Apply CartoonX with mask $s$ and model \(\hat{Q}\) on $x$ to get \(\hat{x}\)
\State Forward \(\hat{x}\) into model \(Q\) and optimize \(Q\)
\EndFor
\EndProcedure
\color{black}
\State For every \(C\) epochs, set parameters \(\hat{Q} \gets Q\)
\Else
\color{gray}
\Procedure{Evaluate model \(Q\)}{}:
\State Project point cloud \(c\) into image \(x\)
\State Apply CartoonX with optimized mask $s$ and model $\hat{Q}$ on image $x$ to get \(\hat{x}\)
\State Forward $\hat{x}$ into model \(Q\) for evaluation
\EndProcedure
\color{black}
\EndIf
% \EndFor
\EndFor 
\end{algorithmic}
\end{algorithm}
%

%===========================================================================================
%
\begin{figure*}[!htb]
    \centering
    \includegraphics[width=0.99\textwidth]{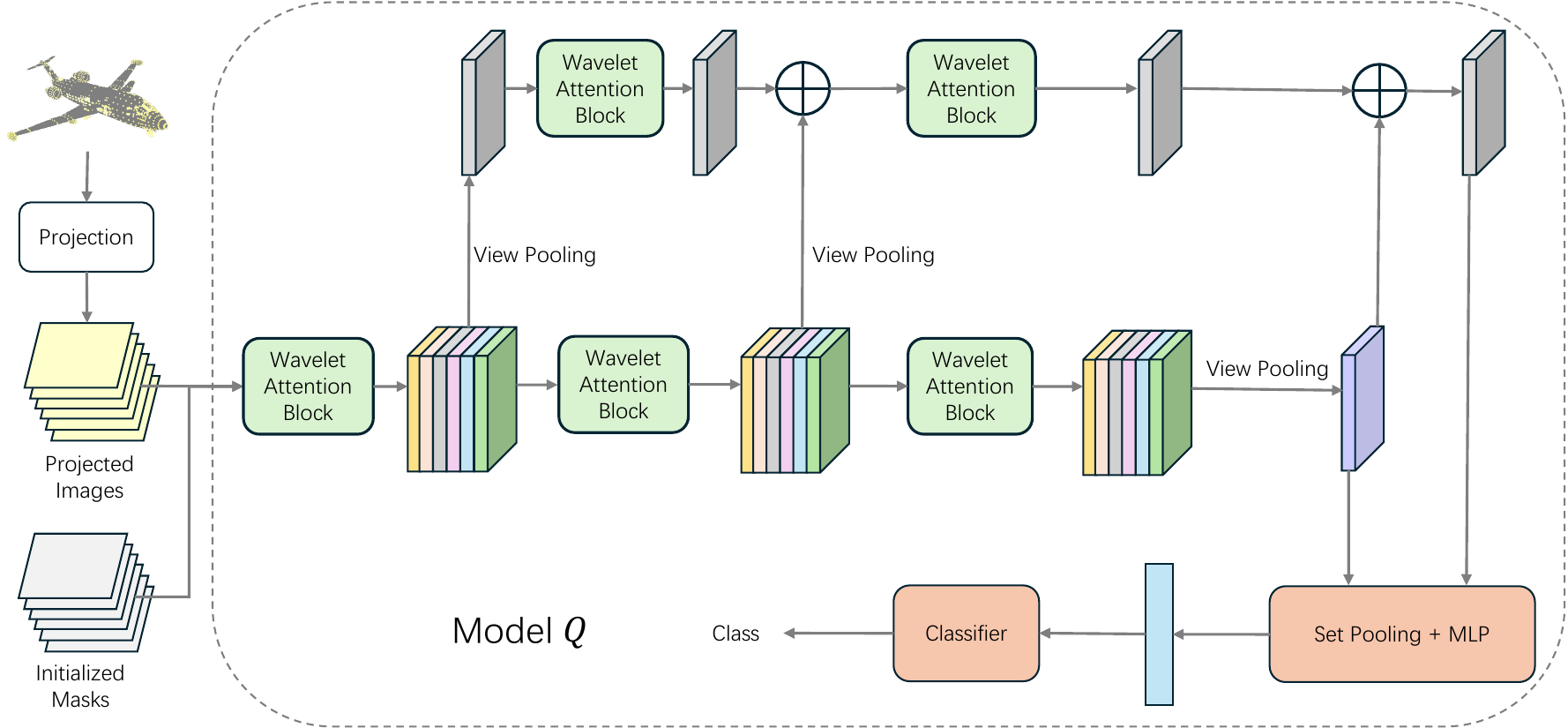}
    \caption{The structure of our RW-Net consisting of model $Q$ (and an identical model $\hat{Q}$) which is built on ViewNet~\cite{chen2023viewnet}. Note that we replace the conventional convolutional block with a wavelet attention block to efficiently capture the semantic contents of the input features while suppressing noise. The symbol $\oplus$ denotes element-wise addition.}
    \label{fig:pipeline}
\end{figure*}

\subsection{Model Architecture}
\label{subsec:arch}
% We begin by extracting features from the projected images using the Wavelet Attention Block\cite{zhao2022wavelet}. Following this, we apply View Pooling\cite{chen2023viewnet} to these projection features to derive initial point features. Subsequently, another Wavelet Attention Block\cite{zhao2022wavelet} processes the initial projection features to generate secondary features. These secondary features are also subjected to View Pooling\cite{chen2023viewnet}, and the resulting view-pooled secondary features are combined with the initially processed point features via a summation operation. This PROCEDURE is iterated to develop the level 3 features. Both the level 3 point features and the view-pooled level 3 projection features are then forwarded to a set pooling layer and a multilayer perceptron (MLP) to synthesize the final feature representation. Once synthesized, these features are input into our few-shot classification head to facilitate the final classification task. The baseline is displayed in Figure \ref{fig:pipeline}

In Figure~\ref{fig:pipeline}, we visualize the internal architecture of model $Q$ (and also $\hat{Q}$ as they are identical). Once we get the masked image $\hat{x}$ using the optimized mask $\hat{s}$ and the projected image $x$ (from six views in total), we apply View Pooling~\cite{chen2023viewnet} to these images to derive the initial pooled images, which are then fed into a Wavelet Attention (WA) Block~\cite{zhao2022wavelet} to extract image features. Meanwhile, another WA Block is applied to process each view of the initial projected images to generate per-view image features, and these per-view features are also pooled with View Pooling. Then, the resulting view-pooled features are fused with the above-mentioned image features via a summation operation. This procedure is iterated two more times to get the wavelet feature and pooled feature. Next, both the wavelet features and the view-pooled features are forwarded to a Set Pooling layer~\cite{mnih2015human} and a multilayer perceptron to get the final image feature. This final feature is fed into a classification head to predict the point cloud class.

In the original ViewNet~\cite{chen2023viewnet}, the view-pooled and per-view images are processed using the conventional Convolutional Neural Network (CNN) blocks. However, we refine this architecture using WA Blocks~\cite{zhao2022wavelet}. As mentioned above, conventional CNNs process a blend of low- and high-frequency components of images, which is not consistent with our human behavior in which we mainly rely on low-frequency components of an image to recognize its content. Therefore, we proposed replacing the conventional CNN blocks in ViewNet with WA Blocks so that the noise contained in the high-frequency components can be suppressed or even filtered out. 
% The mechanism of the wavelet attention block is detailed in the following section.

%===========================================================================================
\subsection{Wavelet Attention Block}
\label{subsec:wavelet}
% The Wavelet-Attention (WA) block\cite{zhao2022wavelet} enhances CNNs by using the Discrete Wavelet Transform (DWT) to decompose feature maps into low-frequency and high-frequency components. The DWT decomposes an input feature map \( X \) into four sub-bands: the low-frequency component \( X_{ll} \) and three high-frequency components \( X_{lh} \) (horizontal detail), \( X_{hl} \) (vertical detail), and \( X_{hh} \) (diagonal detail). This decomposition is mathematically expressed as:

% \resizebox{220pt}{7pt}{$
% X_{ll} = L X L^T, \quad X_{lh} = H X L^T, \quad X_{hl} = L X H^T, \quad X_{hh} = H X H^T
% $}

% where \( L \) and \( H \) are the low-pass and high-pass filter matrices, respectively. The WA block\cite{zhao2022wavelet} discards the \( X_{hh} \) component to reduce noise. The remaining components \( X_{lh} \) and \( X_{hl} \) are combined and normalized using the softmax function to generate an attention map \( \delta \):
% \[
% \delta = \text{softmax}(X_{lh} + X_{hl})
% \]
% This attention map is applied to the low-frequency component \( X_{ll} \) through element-wise multiplication, enhancing detailed features while preserving the main structure. The final output \( Z \) of the WA block\cite{zhao2022wavelet} is:
% \[
% Z = X_{ll} + X_{ll} \odot \delta
% \]
% where \( \odot \) denotes element-wise multiplication.

\begin{figure}[!htb]
    \centering
    \includegraphics[width=0.99\linewidth]{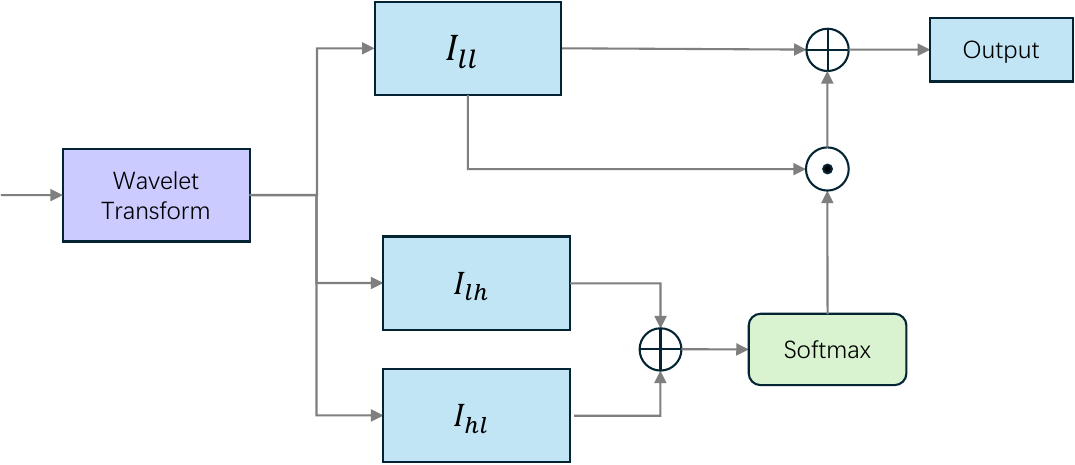}
    \caption{Structure of wavelet attention block. $\mathbf{I}_{ll}$, $\mathbf{I}_{lh}$, and $\mathbf{I}_{hl}$ adapted from~\cite{zhao2022wavelet}. The symbol $\oplus$ denotes element-wise addition, and $\odot$ denotes element-wise multiplication.}
    \label{fig:wavelet}
\end{figure}

We adopt the Wavelet Attention Block proposed in~\cite{zhao2022wavelet}, which enhances conventional CNNs by first using the Discrete Wavelet Transform (DWT) to decompose an image feature map into low-frequency and high-frequency components, and then fusing them with attention~\cite{vaswani2017attention}. Specifically, the DWT decomposes an input feature map $\textbf{I}$ into four sub-bands: the low-frequency component $\mathbf{I}_{ll}$ and three high-frequency components $\mathbf{I}_{lh}$ (horizontal detail), $\mathbf{I}_{hl}$  (vertical detail), and $\mathbf{I}_{hh}$  (diagonal detail), which is mathematically expressed in Eq.~\ref{eq:decompose}. The WA block discards the $\mathbf{I}_{hh}$ component to reduce noise. The remaining components $\mathbf{I}_{lh}$ and $\mathbf{I}_{hl}$ are combined and normalized using the softmax function to generate an attention map $\Delta_{att}$:
\begin{equation}
\Delta_{att} = \text{softmax}(\mathbf{I}_{lh} + \mathbf{I}_{hl})\enspace.
\end{equation}
This attention map is then fused with the low-frequency component $\mathbf{I}_{ll}$ through element-wise multiplication, enhancing detailed features while preserving the main structure. The final output of the WA block\cite{zhao2022wavelet} is:
\begin{equation}
\textbf{Z} = \mathbf{I}_{ll}+ \mathbf{I}_{ll} \odot \Delta_{att}\enspace,
\end{equation}
where \( \odot \) denotes element-wise multiplication. The illustration of the internal structure of a WA block is shown in Figure~\ref{fig:wavelet}.

%

%%%%%%%%%%%%%%%%%%%%%%%%%%%%%%%%%%%%%%%%%%%%%%%%%%%%%%%%%%%%%%%%%%%%%%%%%%%%%%%%%%%%%%%%%%%%%%%%%%%%%%%%%%
\section{Experiments}
\label{sec:exp}
In this section, we first describe the experiment setup in Section~\ref{subsec:setup}, including datasets on which we train and evaluate our model, and all the baselines we compare with. Then, in the following Section~\ref{subsec:scan} to Section~\ref{subsec:vis}, we show our quantitative results and qualitative visualizations.

%===========================================================================================
\subsection{Setup}
\label{subsec:setup}
\noindent \textbf{Datasets.} ModelNet40~\cite{wu20153d} includes 12,311 3D CAD models across 40 object classes, extensively used for training and evaluating methods for 3D shape recognition, object classification, and part segmentation. ModelNet40-C\cite{sun2022benchmarking} is an extension of ModelNet40, designed to benchmark the robustness of 3D point cloud recognition models against common corruptions, featuring 185,100 point clouds across 40 classes with 15 types of corruption and five severity levels. ScanObjectNN\cite{uy2019revisiting} includes real-world scanned 3D point clouds, rather than synthetic ones, presenting additional challenges like background clutter, occlusions, and incomplete data.

\noindent \textbf{Training.}
For the pre-training process, we trained the network using the Adam optimizer \cite{kingma2014adam} with a learning rate of 0.0001 over 20 epochs without applying CartoonX to the model. During the training phase, we continued to use the Adam optimizer at the same learning rate for 60 epochs, updating the pre-trained model every 20 epochs. For CartoonX\cite{kolek2022cartoon} optimization, we set \(\lambda\) to 0.01, adjusted the learning rate to 0.1, and applied Gaussian obfuscation. We initialized the mask as all one and optimized for 30 epochs. For other baseline methods, we employed the Adam optimizer with a learning rate of 0.0001 across 80 epochs.

\noindent \textbf{Evaluation.}
We evaluate our model based on the mean accuracy achieved on all episodes in multiple folds, with a confidence interval of 95\%.

\noindent \textbf{Baselines.}
To verify the effectiveness of our method, we selected four state-of-the-art baselines for comparison, including DGCNN\cite{wang2019dynamic}, 
GPr-Net\cite{anvekar2023gpr},
ViewNet\cite{chen2023viewnet}, 
SimpliMix\cite{yang2024SimpliMix} with DGCNN\cite{wang2019dynamic}, and SimpliMix\cite{yang2024SimpliMix} with ViewNet\cite{chen2023viewnet}. For the DGCNN\cite{wang2019dynamic} backbone, we employed CIA\cite{ye2022makes} and Protonet\cite{snell2017prototypical} as the few-shot head, for GPr-Net\cite{anvekar2023gpr} we used ProtoNet\cite{snell2017prototypical} as few-shot head, and for other 
backbones we applied CIA\cite{ye2022makes} as the few-shot head.

%===========================================================================================
\subsection{ScanObjectNN Results}
\label{subsec:scan}
For the ScanObjectNN\cite{uy2019revisiting} dataset, we conducted 5-way 1-shot 10-query and 5-way 5-shot 10-query classifications. For cross-validation, we divided the dataset into three equal-sized folds. The results, presented in Table \ref{tab:scanobj}, demonstrate that our approach outperforms the others in both 1-shot and 5-shot classifications. Specifically, for the 1-shot classification, our method increased the mean accuracy by 0.58\% compared to the best baseline, \ie, ViewNet\cite{chen2023viewnet}. Similarly, for the 5-shot classification, our method improves the mean accuracy by 0.64\%. Also note that, on average, our RW-Net achieves smaller standard derivations, which indicates the stability of our method.
\begin{table}[ht]
\centering
\scriptsize
\setlength{\tabcolsep}{2.2pt}
% \resizebox{230pt}{70pt}{
\caption{Comparison between the performance of baselines and our model on ScanObjectNN. The best performance is in bold, and the second best is underlined (same in all tables).}
\begin{tabular}{lcccc}
\toprule
& \textbf{Fold 0} & \textbf{Fold 1} & \textbf{Fold 2} & \textbf{Mean}   \\
\midrule
\multicolumn{5}{c}{\textbf{5-way 1-shot}} \\
ProtoNet (DGCNN)    & 48.68±0.54\% & 61.48±0.63\% & 60.38±0.19\% & 56.85±0.45\% \\
DGCNN (CIA)         & 49.47±0.82\% & 63.28±0.76\% & 61.79±0.72\% & 58.18±0.77\% \\
GPr-Net         & 51.37±0.81\% & 62.74±0.80\% & 61.95±0.69\% &  58.69±0.77\% \\
ViewNet         & \underline{60.29±0.63\%} & \underline{63.53±0.86\%} & 62.10±0.72\% & \underline{61.97±0.73\%} \\
SimpliMix with DGCNN         & 54.36±0.57\% & 60.26±0.59\% & 59.41±0.48\% & 59.01±0.55\% \\
SimpliMix with ViewNet         & 59.98±0.77\% & 63.18±0.47\% & \underline{62.28±0.52\%} & 61.81±0.58\% \\
Ours            & \textbf{60.84±0.45\%} & \textbf{63.92±0.52\%} & \textbf{62.89±0.51\%} & \textbf{62.55±0.49\%} \\
\midrule
\multicolumn{5}{c}{\textbf{5-way 5-shot}} \\
DGCNN (ProtoNet)        & 61.32±0.87\% & 67.91±0.49\% & 68.53±0.46\% & 65.92±0.61\% \\
DGCNN (CIA)         & 62.59±0.62\% & 68.32±0.55\% & 68.92±0.51\% & 66.61±0.56\% \\
GPr-Net         & 63.45±0.82\% & 71.59±0.57\% & 75.32±0.49\% &  70.12±0.63\% \\
ViewNet         & 70.80±0.78\% & \textbf{74.42±0.82\%} & \underline{76.89±0.76\%} & \underline{74.04±0.78\%} \\
SimpliMix with DGCNN         & 65.45±0.31\% & 71.18±0.37\% & 73.31±0.56\% & 69.98±0.41\% \\
SimpliMix with ViewNet         & \underline{71.38±0.93\%} & 70.63±0.70\% & 75.19±0.72\% & 72.82±0.78\% \\
Ours            & \textbf{72.96±0.51\%} & \underline{73.95±0.69\%}\ & \textbf{77.13±0.58\%} & \textbf{74.68±0.59\%} \\
\bottomrule
\end{tabular}
% }
\label{tab:scanobj}
\end{table}
%

%===========================================================================================
\subsection{ModelNet40 Results}
\label{subsec:modelnet40}
Similarly to the settings for ScanObjectNN\cite{uy2019revisiting} dataset, we conducted 5-way 1-shot 10-query and 5-way 5-shot 10-query classifications on the ModelNet40 dataset. For ModelNet40, we divided the dataset into four equal folds. The test results, shown in Table \ref{tab:modelnet}, indicate that our method outperforms the baselines in all folds. Specifically, for the 1-shot classification, our method achieves a 1.3\% increase in mean accuracy against SimpliMix with ViewNet, and for the 5-shot classification, it surpasses the baseline by an average of 1.02\% across all folds. From Table~\ref{tab:scanobj} and Table~\ref{tab:modelnet}, we can see that our RW-Net performs well on both synthetically intact and real scanned incomplete 3D objects.
\begin{table}[ht]
\centering
\tiny
\caption{Comparison between the performance of baselines and our model on ModelNet40.}
\setlength{\tabcolsep}{4pt}
% \resizebox{230pt}{70pt}{
\centering
\begin{tabular}{lccccc}
\toprule
& \textbf{Fold 0} & \textbf{Fold 1} & \textbf{Fold 0} & \textbf{Fold 3} & \textbf{Mean}   \\
\midrule
\multicolumn{6}{c}{\textbf{5-way 1-shot}} \\
DGCNN (ProtoNet)         & 87.93±0.43\% & 82.85±0.46\% & 73.09±0.74\% & 74.67±0.67\% & 79.64±0.58\% \\
DGCNN (CIA)         & 89.05±0.31\% & \textbf{83.12±0.77}\% & 73.15±0.85\% & 75.20±0.52\% & 80.13±0.61\% \\
GPr-Net         & 80.95±0.47\% & 81.60±0.54\% & 71.15±0.85\% & 70.20±0.52\% & 75.98±0.60\% \\
ViewNet         & 91.80±0.57\% & 81.01±0.72\% & \underline{75.32±0.61\%} & \underline{80.67±0.68\%} & 82.20±0.65\% \\
SimpliMix with DGCNN         & 89.29±0.26\% & 81.57±0.31\% & 73.48±0.45\% & 76.71±0.34\% & 80.26±0.34\% \\
SimpliMix with ViewNet         & \underline{92.21±0.46\%} & 81.75±0.39\% & 75.04±0.41\% & 80.08±0.50\% & \underline{82.27±0.44\%} \\
Ours            & \textbf{93.43±0.34\%} & \underline{82.47±0.47\%} & \textbf{77.09±0.29\%} & \textbf{81.28±0.38\%} & \textbf{83.57±0.37\%} \\
\midrule
\multicolumn{6}{c}{\textbf{5-way 5-shot}} \\
DGCNN (ProtoNet)         & 93.72±0.46\% & 88.18±0.47\% & 84.48±0.60\% & 84.32±0.45\% & 87.68±0.49\% \\
DGCNN (CIA)         & 93.51±0.31\% & 89.04±0.52\% & 85.23±0.75\% & 85.91±0.31\% & 88.42±0.47\% \\
GPr-Net         & 83.70±0.53\% & 78.31±0.67\% & 75.39±0.82\% & 71.30±0.87\% & 77.18±0.72\% \\
ViewNet         & \textbf{96.06±0.43\%} & 89.20±0.51\% & 85.73±0.66\% & \underline{88.91±0.49\%} & \underline{89.98±0.52\%} \\
SimpliMix with DGCNN         & 94.01±0.36\% & \underline{89.21±0.41\%} & 85.39±0.45\% & 87.28±0.71\% & 88.97±0.48\% \\
SimpliMix with ViewNet         & 95.32±0.37\% & 88.97±0.61\% & \underline{86.12±0.70\%} & 88.13±0.52\% & 89.64±0.55\% \\
Ours            & \underline{95.47±0.53\%} & \textbf{90.20±0.57\%} & \textbf{86.69±0.39\%} & \textbf{91.03±0.42\%} & \textbf{90.85±0.48\%} \\
\bottomrule
\end{tabular}
% }
\label{tab:modelnet}
\end{table}
%

%===========================================================================================
\subsection{ModelNet40-C Results}
\label{subsec:modelnet40c}
For ModelNet40-C\cite{sun2022benchmarking}, we adopted the same settings as used for ModelNet40 and ScanObjectNN. The experimental results are summarized in Table \ref{tab:modelnet_c}. Our method outperforms the baselines in all folds. Specifically, in the 1-shot classification, our method surpasses the best baseline, \ie, SimpliMix with ViewNet, by 1.35\%, and in the 5-shot classification, it exceeds by 1.28\%. Also, notice that our RW-Net outperforms the best baseline across all folds. From Table~\ref{tab:scanobj} and Table~\ref{tab:modelnet_c}, we conclude that our method is robust against both real-world shape artifacts and man-made corruptions.
\begin{table}[ht]
\centering
\tiny
\setlength{\tabcolsep}{4pt}
\caption{Comparison between the performance of baselines and our model on ModelNet40-C.}
% \resizebox{230pt}{70pt}{
\centering
\begin{tabular}{lccccc}
\toprule
& \textbf{Fold 0} & \textbf{Fold 1} & \textbf{Fold 0} & \textbf{Fold 3} & \textbf{Mean}   \\
\midrule
\multicolumn{6}{c}{\textbf{5-way 1-shot}} \\
DGCNN (ProtoNet)         & 80.80±0.70\% & 79.04±0.72\% & 63.18±0.79\% & 72.91±0.65\% & 73.98±0.72\% \\
DGCNN (CIA)         & 84.13±0.89\% & 79.22±0.75\% & 64.15±0.85\% & 73.30±0.71\% & 75.20±0.80\% \\
GPr-Net         & 70.25±0.49\% & 68.40±0.31\% & 60.35±0.70\% & 65.27±0.39\% & 66.07±0.47\% \\
ViewNet         & \underline{89.13±0.73\%} & 80.01±0.64\% & \underline{70.71±0.66\%} & 74.50±0.68\% & 78.59±0.68\% \\
SimpliMix with DGCNN         & 85.38±0.42\% & \underline{80.51±0.31\%} & 68.35±0.21\% & 72.67±0.73\% & 76.73±0.42\% \\
SimpliMix with ViewNet         & 88.78±0.67\% & 80.27±0.72\% & 70.40±0.81\% & \textbf{76.67±0.45\%} & \underline{79.03±0.66\%} \\
Ours            & \textbf{91.53±0.42\%} & \textbf{82.01±0.51\%} & \textbf{71.91±0.29\%} & \underline{76.05±0.38\%} & \textbf{80.38±0.40\%} \\
\midrule
\multicolumn{6}{c}{\textbf{5-way 5-shot}} \\
DGCNN (ProtoNet)        & 92.16±0.69\% & 83.01±0.70\% & 77.92±0.74\% & 80.09±0.58\% & 83.30±0.68\% \\
DGCNN (CIA)         & 92.28±0.52\% & 83.22±0.75\% & 79.15±0.85\% & 80.30±0.71\% & 83.74±0.71\% \\
GPr-Net         & 72.14±0.51\% & 71.39±0.48\% & 65.39±0.81\% & 67.29±0.88\% & 69.05±0.67\% \\
ViewNet         & 93.85±0.31\% & 87.08±0.39\% & 80.25±0.47\% & 87.40±0.51\% & 87.15±0.42\% \\
SimpliMix with DGCNN         & 93.15±0.49\% & 82.17±0.71\% & 80.48±0.21\% & 80.71±0.54\% & 84.13±0.49\% \\
SimpliMix with ViewNet      & \underline{95.19±0.43\%} & \underline{87.71±0.37\%} & \underline{82.63±0.28\%} & \underline{87.85±0.33\%} & \underline{88.11±0.35\%} \\
Ours            & \textbf{95.97±0.71\%} & \textbf{88.90±0.63\%} & \textbf{82.79±0.54\%} & \textbf{89.91±0.61\%} & \textbf{89.39±0.53\%} \\
\bottomrule
\end{tabular}
% }
\label{tab:modelnet_c}
\end{table}
%

%===========================================================================================
\subsection{Visualization of Model Components}
\label{subsec:vis}
\begin{figure}[!htb]
\centering
\includegraphics[width=0.99\linewidth]{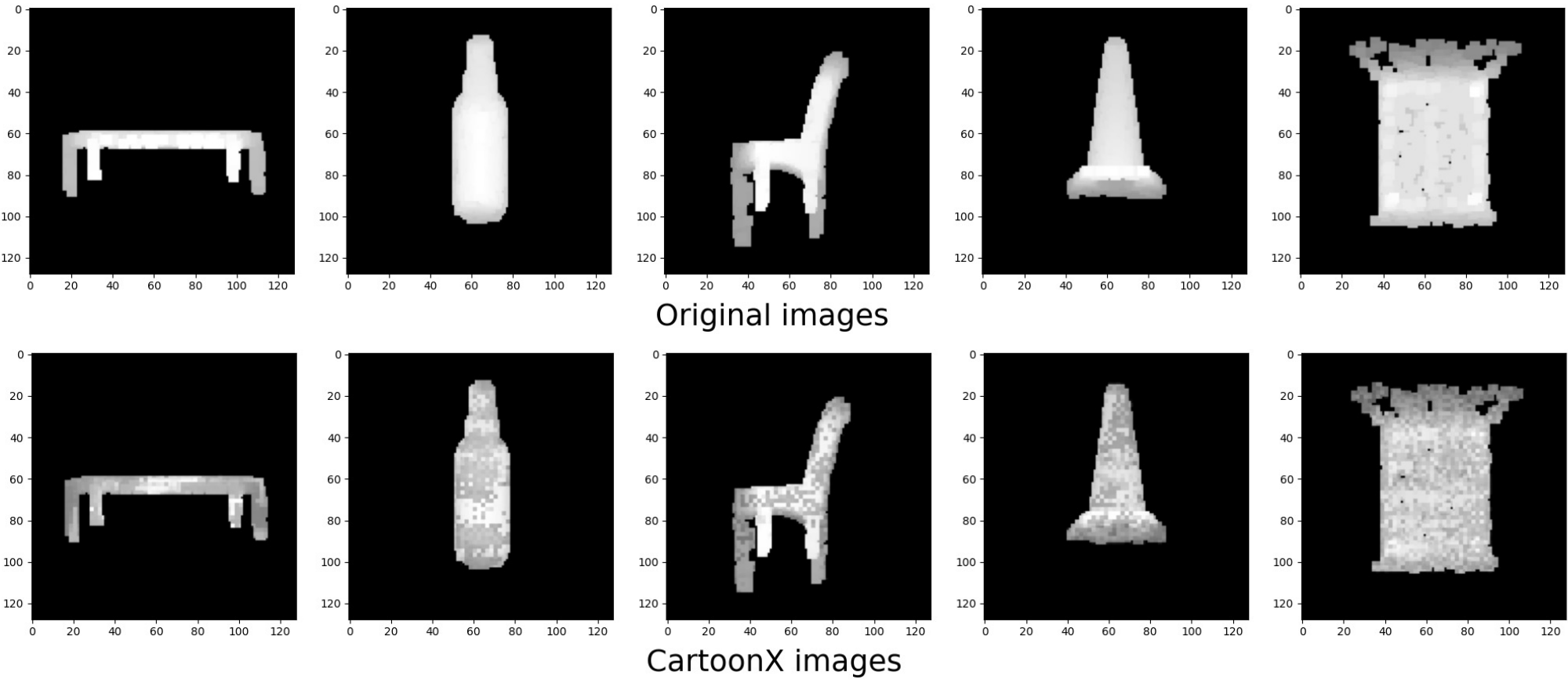}
\caption{Comparison between unprocessed projected images (top) and cartoonX processed images (bottom).}
\label{fig:cartoonX}
\end{figure}

\begin{figure}[!htb]
\includegraphics[width=0.99\linewidth]{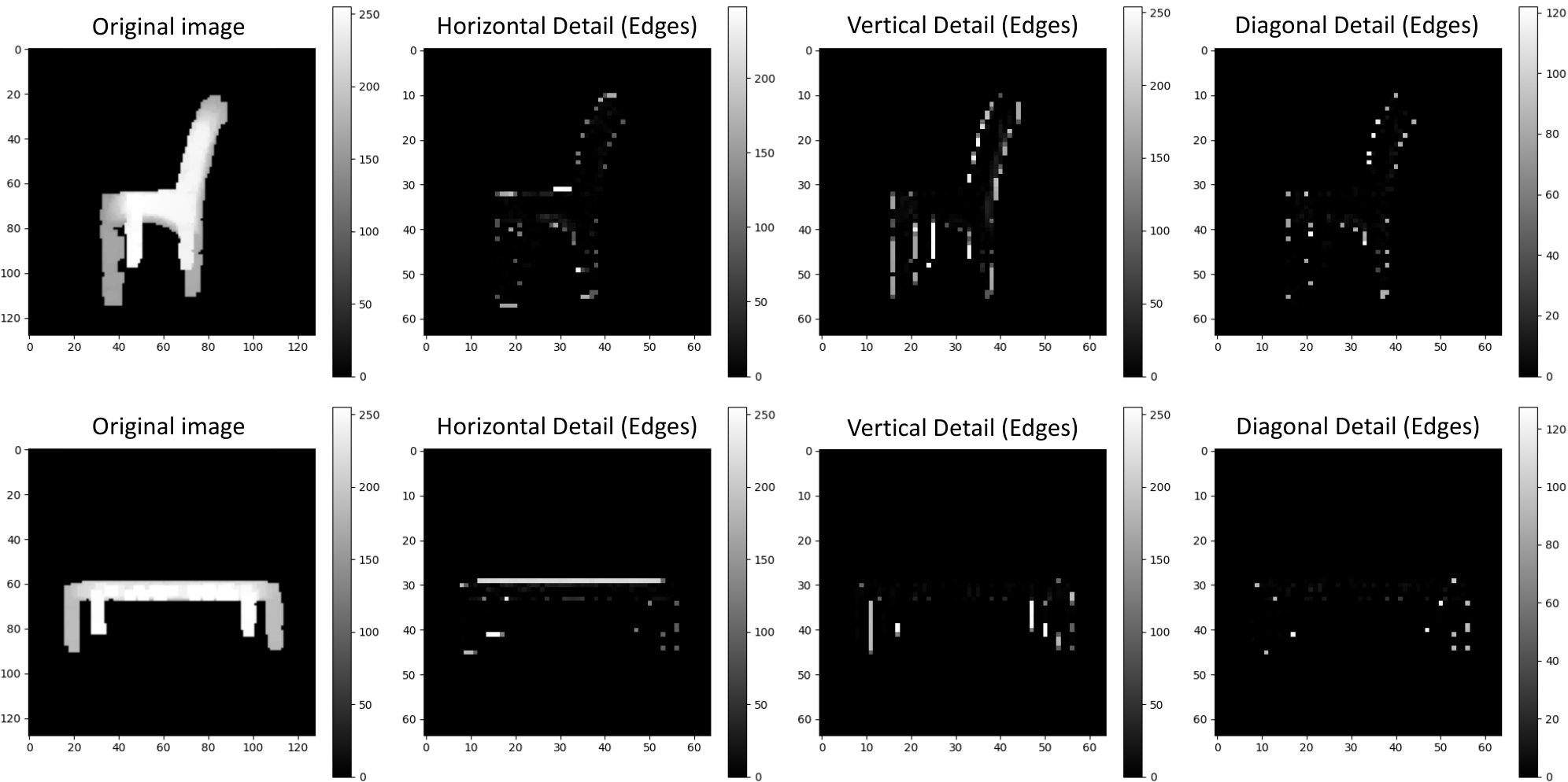}
\caption{Visualization of wavelet transform extracting different frequency components of the original images.}
\label{fig:vis}        
\end{figure}

To gain a more comprehensive understanding of the functionality and behavior of our proposed model, we performed a detailed visualization analysis on a subset of the projected images. Specifically, Figure \ref{fig:cartoonX} demonstrates the effect of the CartoonX procedure, which is designed to selectively obscure or emphasize certain regions of the original model's projection. In addition, Figure \ref{fig:vis} shows the application of the wavelet transform. The wavelet transform is shown to effectively decompose the projected images into different frequency components, isolating low-frequency features that capture the fundamental geometric structure and high-frequency features that encode fine-grained details and possible noise.

%
% \begin{figure*}[!htb]
%     \centering
%     \begin{subfigure}[b]{0.45\textwidth}
%         \includegraphics[width=0.99\textwidth]{figures/cartoonX.jpg}
%         \caption{Comparison between unprocessed projected images and cartoonX processed images}
%         \label{fig:cartoonX}
%     \end{subfigure}
%     \begin{subfigure}[b]{0.45\textwidth}
%         \includegraphics[width=0.99\textwidth]{figures/Visualization.jpg}
%         \caption{Visualization of wavelet transform extracting different features from original images}
%         \label{fig:Visualization}        
%     \end{subfigure}
% \end{figure*}
%

%

%%%%%%%%%%%%%%%%%%%%%%%%%%%%%%%%%%%%%%%%%%%%%%%%%%%%%%%%%%%%%%%%%%%%%%%%%%%%%%%%%%%%%%%%%%%%5
\section{Ablation Study}
\label{sec:abl}
In our ablation studies, we use the ScanObjectNN\cite{uy2019revisiting} dataset to analyze the effectiveness of different settings and components of our model.

%===========================================================================================
\subsection{Analysis of Bin-wise Loss}
Unlike the DGCNN\cite{wang2019dynamic} backbone, which produces a feature vector \( f \in \mathbb{R}^d \) where \( d \) denotes vector dimension to feed into the few-shot head, our method, similar to ViewNet\cite{chen2023viewnet}, generates an output \( O \in \mathbb{R}^{b \times d} \), where \( b \) represents the number of bins. For this ablation study, we conducted a bin-wise loss comparison, introducing a linear layer to compress \( O \) into \( \tilde{O} \), which matches the dimensions of \( f \). We performed 5-way 1-shot and 5-way 5-shot classifications on both feature vectors. According to the results in Table \ref{Bin}, there is a noticeable drop in accuracy using the compression. This also possibly and partially explains why DGCNN performs worse, on average, in Table~\ref{tab:scanobj} to Table~\ref{tab:modelnet_c}.
\begin{table}[ht]
\centering
\footnotesize
% \resizebox{180pt}{40pt}{
\centering
\caption{Comparison between compression output $\tilde{O}$ and bin-wise $O$.}
\begin{tabular}{lcccc}
\toprule
& \textbf{Fold 0} & \textbf{Fold 1} & \textbf{Fold 2} & \textbf{Mean}   \\
\midrule
\multicolumn{5}{c}{\textbf{5-way 1-shot}} \\
$\tilde{O}$              & 58.27\% & 62.19\% & 59.25\% & 59.90\% \\
$O$               & \textbf{60.84\%} & \textbf{63.92\%} & \textbf{62.89 \%} & \textbf{62.55\%} \\
\midrule
\multicolumn{5}{c}{\textbf{5-way 5-shot}} \\
$\tilde{O}$       & 68.22\% & 73.43\% & 74.96\% & 72.20\% \\
$O$               & \textbf{72.96\%} & \textbf{73.95\%} & \textbf{77.13\%} & \textbf{74.68\%} \\
\bottomrule
\end{tabular}
% }
\label{Bin}
\end{table}
%

%===========================================================================================
\subsection{Study on Pre-trained Model and Final Model}
As mentioned above, we initially developed a pre-training model, denoted as \( \hat{Q} \), which served as the precursor to our final model \( Q \). After completing the pre-training phase, we utilize the pre-trained model \( \hat{Q} \) to enhance the training of model \( Q \). To quantitatively evaluate the efficacy of this strategy, we compared the highest accuracy achieved by the pre-trained model \( \hat{Q} \) against that of the final model \( Q \). The results in Table \ref{Pre} indicate a significant improvement in accuracy from model \( \hat{Q} \) to model \( Q \). We attribute this improvement to the strategy of iteratively updating the frozen model $\hat{Q}$ with the parameters of model $Q$ while continuously optimizing model $Q$ with the mask $\hat{s}$ produced by the frozen model $\hat{Q}$.
\begin{table}[ht]
\centering
% \resizebox{180pt}{30pt}{
\footnotesize
\centering
\caption{Comparison between pre-trained model and final model.}
\begin{tabular}{lcccc}
\toprule
& \textbf{Fold 0} & \textbf{Fold 1} & \textbf{Fold 2} & \textbf{Mean}   \\
\midrule
\multicolumn{5}{c}{\textbf{5-way 1-shot}} \\
Pre-trained $\hat{Q}$        & 58.67\% & 62.29\% & 61.59\% & 60.85\% \\
Final $Q$            & \textbf{60.84\%} & \textbf{63.92\%} & \textbf{62.89\%} & \textbf{62.55\%} \\
\midrule
\multicolumn{5}{c}{\textbf{5-way 5-shot}} \\
Pre-trained $\hat{Q}$        & 71.98\% & 73.12\% & 76.29\% & 73.80\% \\
Final $Q$           & \textbf{72.96\%} & \textbf{73.95\%} & \textbf{77.13\%} & \textbf{74.68\%} \\
\bottomrule
\end{tabular}
% }
\label{Pre}
\end{table}
%

%===========================================================================================
\subsection{Study on Effectiveness of Wavelet Attention Block}
To demonstrate the efficacy of the Wavelet Attention (WA) Block \cite{zhao2022wavelet} within our model, we substitute all max-pooling blocks in ViewNet \cite{chen2023viewnet} with the WA Block \cite{zhao2022wavelet}. The experimental results, shown in Table \ref{WABlock}, confirm that replacing max-pooling blocks with the WA block \cite{zhao2022wavelet} enhances the performance of ViewNet \cite{chen2023viewnet}. This confirms that wavelet transform is able to retain geometric and structural features of 3D objects, while max-pooling mixes all features together and impedes the process of extracting meaningful shape features. 
\begin{table}[ht]
\centering
% \resizebox{180pt}{30pt}{
\footnotesize
\centering
\caption{Comparison between max-pooling and WA block Substituting max-pooling in ViewNet.}
\begin{tabular}{lcccc}
\toprule
& \textbf{Fold 0} & \textbf{Fold 1} & \textbf{Fold 2} & \textbf{Mean}   \\
\midrule
\multicolumn{5}{c}{\textbf{5-way 1-shot}} \\
Max-pooling        & 60.29\% & 63.53\% & 62.10\% & 61.97\% \\
WA Block       & \textbf{60.59\%} & \textbf{63.74\%} & \textbf{62.36\%} & \textbf{62.23\%} \\
\midrule
\multicolumn{5}{c}{\textbf{5-way 5-shot}} \\
Max-pooling        & 70.80\% & \textbf{74.42\%} & 76.89\% & 74.04±0.78\% \\
WA Block            & \textbf{72.02\%} & 74.02\% & \textbf{77.01\%} & \textbf{74.35\%} \\
\bottomrule
\end{tabular}
% }
\label{WABlock}
\end{table}

To thoroughly evaluate the effectiveness of each component in the Wavelet Attention Block\cite{zhao2022wavelet}, 
% which incorporates three out of the four decompositions from the Discrete Wavelet Transform, specifically \( \mathbf{I}_{ll} \), \( \mathbf{I}_{lh} \), and \( \mathbf{I}_{hl} \), we conducted a series of experiments.
we replaced the \( \mathbf{I}_{ll} \) component with \( \mathbf{I}_{hh} \) to assess the impact of substituting low-frequency information with high-frequency information. We also separately removed the \( \mathbf{I}_{lh} \) and \( \mathbf{I}_{hl} \) components to determine their contributions to the WA Block’s\cite{zhao2022wavelet} performance.
The experiment results in Table \ref{tab:WABlock} indicate that the original version of the WA Block\cite{zhao2022wavelet} achieves the best performance. When the \( \mathbf{I}_{ll} \) component is replaced with \( \mathbf{I}_{hh} \), there is a significant reduction in accuracy. This drop is attributed to the \( \mathbf{I}_{hh} \) component, which contains fine-grained geometric details but also contains too much noise from the input, adversely affecting the performance. Furthermore, removing the blocks with \( \mathbf{I}_{lh} \) and \( \mathbf{I}_{hl} \) components both exhibit lower accuracy. This decrease is likely due to the loss of fine-grained information from \( \mathbf{I}_{lh} \) and \( \mathbf{I}_{hl} \). 
\begin{table}[ht]
\centering
\footnotesize
% \resizebox{180pt}{30pt}{
\caption{Comparison of different versions of WA Block.}
\begin{tabular}{lcccc}
\toprule
& \textbf{Fold 0} & \textbf{Fold 1} & \textbf{Fold 2} & \textbf{Mean}   \\
\midrule
\multicolumn{5}{c}{\textbf{5-way 1-shot}} \\
Original model       & \textbf{60.84\%} & \textbf{63.92\%} & \textbf{62.89 \%} & \textbf{62.55\%} \\
\(I_{hh}\) replaced            & 58.35\% & 61.17\% & 61.01 \% & 60.51\% \\
\(I_{lh}\) removed            & 59.31\% & 62.29\% & 61.97 \% & 61.19\% \\
\(I_{hl}\) removed            & 58.96\% & 62.43\% & 62.07 \% & 61.15\% \\
\bottomrule
\end{tabular}
% }
\label{tab:WABlock}
\end{table}
%

%===========================================================================================
\subsection{Analysis on Different Settings of CartoonX}
We investigate the impact of various CartoonX\cite{kolek2022cartoon} settings on our model's performance. To achieve this, we maintained all other parameters constant while systematically varying the \(\lambda\) value in Eq.~\ref{eq:opt_cont}. The results of these experiments are presented in Table~\ref{tab:CartoonX}. We observe a noticeable drop in accuracy as the \(\lambda\) value increases. This trend suggests that higher \(\lambda\) values lead to a much sparse mask $\hat{s}$, thus resulting in the excessive removal of essential information from the images, which in turn hampers the model's ability to accurately classify and recognize objects.
\begin{table}[ht]
\centering
\footnotesize
% \resizebox{180pt}{30pt}{
\caption{Comparison of different \(\lambda\) in CartoonX.}
\begin{tabular}{lcccc}
\toprule
& \textbf{Fold 0} & \textbf{Fold 1} & \textbf{Fold 2} & \textbf{Mean}   \\
\midrule
\multicolumn{5}{c}{\textbf{5-way 1-shot}} \\
\(\lambda = 0.01\)       & \textbf{60.84\%} & \textbf{63.92\%} & \textbf{62.89} \% & \textbf{62.55\%} \\
\(\lambda = 1\)             & 60.35\% & 63.57\% & 62.71 \% & 62.21\% \\
\(\lambda = 10\)            & 59.91\% & 63.08\% & 61.47 \% & 61.49\% \\
\(\lambda = 100\)            & 58.42\% & 60.78\% & 60.03 \% & 59.74\% \\
\bottomrule
\end{tabular}
% }
\label{tab:CartoonX}
\end{table}
%

%===========================================================================================
\subsection{Analysis on Computational Cost}
We acknowledge that adding RDE and Wavelet Attention Blocks indeed increases the computation cost, as shown in Table~\ref{tab:cost}: forward/backward size refers to the memory required to store activations during the forward pass and to compute gradients during the backward pass, and parameter size refers to the memory size occupied by the learnable parameters (such as weights and biases) of the model. However, it also brings performance gains on all three datasets over all the compared baselines, as evidenced by Table~\ref{tab:scanobj} to Table~\ref{tab:modelnet_c}.

\begin{table}[ht]
\centering
% \resizebox{180pt}{20pt}{
\footnotesize
\caption{Computational cost between ViewNet and our RW-Net.}
\begin{tabular}{lcccc}
\toprule
& Forw./Backw. Pass Size & Param Size  & Total Size   \\
\midrule
ViewNet        & 106.50 MB &  6.70 MB & 113.22 MB \\
RW-Net       & 217.15 MB &  14.90 MB & 232.05 MB  \\
\bottomrule
\end{tabular}
% }
\label{tab:cost}
\end{table}
%

%%%%%%%%%%%%%%%%%%%%%%%%%%%%%%%%%%%%%%%%%%%%%%%%%%%%%%%%%%%%%%%%%%%%%%%%%%%%%%%%%%%%%%%%%%%%5
\section{Conclusion}
\label{sec:con}
We have developed a robust framework, RW-Net, for few-shot 3D point cloud classification, leveraging the integration of Rate-Distortion Explanation and wavelet transform within a projection-based model. This innovative integration enables our framework to effectively balance feature extraction and generalization, addressing key challenges in few-shot learning. Our experimental results demonstrate that the proposed approach achieves superior performance and enhanced generalization across multiple benchmark datasets, including ModelNet40, ModelNet40-C, and ScanObjectNN. In the future, we aim to further explore how explanatory methods, such as Rate-Distortion Explanation, can be harnessed not only to develop 3D object classification models, but also to inspire the design of novel models for broader 3D vision applications, such as segmentation and generation. We believe this line of research has significant potential to advance the field by combining explainability and interpretability with network optimization in 3D point cloud analysis.

{\small
\bibliographystyle{IEEEtran}
\bibliography{ref.bib}
}

%%%%%%%%%%%%%%%%%%%%%%%%%%%%%%%%%%%%%%%%%%%%%%%%%%%%%%%%%%%%%%%%%%%%%%%%%%%%%%%%%%%%%%%%%%%%%
% \section*{Appendix}
% \label{sec:app}
%
\begin{algorithm*}
\caption{Full pseudocode for RW-Net training.}
\label{alg:full}
% \small
\begin{algorithmic}[1]
\State Initialize pre-training epoch $M^\prime$, training epoch $M$, CartoonX sparsity $\lambda$, CartoonX optimization step $P$, CartoonX measure of distortion $d$, parameter copying step $C$.
\State Initialize classification model \(Q\) and target model \(\hat{Q}\) with the same structure.  
\State Initialize dictionary \(S\) to store mask $s$ and $\hat{s}$.
\For{$i\gets 1$ to $M^\prime$ epochs}
\Procedure{Pre-train model \(\hat{Q}\)}{}:
\State Project point cloud  \(c\) into image \(x\)
\State Forward \(x\) into  model \(\hat{Q}\)
\EndProcedure
\EndFor
\State Copy parameters \(Q \gets \hat{Q}\)
% \red{\Comment{The relation between $Q$ and $\hat{Q}$ is unclear.}}
\For{$i\gets 1$ to $M$ epochs} 
% \For{$j\gets 1$ to $N$ batches} 
\If{training model}
\Procedure{train model \(Q\)}{}:
\For{$j\gets 1$ to $N$ batches} 
\State Project point cloud \(c\) into image \(x\)
\Procedure{CartoonX}{}:
\If{i == 1}
\State Initialize mask $s \gets \mathbf{1}_{x.\text{shape}}$
\Else
\State Get mask $s = S(x)$
\EndIf
\State Get DWT coefficients \(h\) with \(x=f(h)\) where \(f\) is inverse DWT
\For{$k\gets 1$ to $P$} 
\State Produce adaptive Gaussian noise \(v\)
\State Compute obfuscation \(y\) with \(y:=f(s\odot h + (1-s) \odot v)\)
\State Clip \(y\) into the range of \([0,1]\)
\State Compute distortion \(\hat{D}(h, s,\hat{Q}):=d(\hat{Q}(x),\hat{Q}(y))^2\)
\State Compute loss for mask \(\ell(s) = \hat{D}(h,s,\hat{Q}) + \lambda \Vert s \Vert_1 \) 
\State Update mask $s$ using gradient descent \(s \leftarrow s - \nabla_s \ell(s)\) 
\State Clip $s$ into the range of \([0,1]\)
\EndFor
\State Assign $\hat{s} \leftarrow s$
\State Update $S$ with $S(x)= \hat{s}$
\State return mask \(\hat{s}\)
\EndProcedure
\State Get DWT coefficients $h$ with $x = f(h)$
\State Set masked image \(\hat{x} := f(h \odot \hat{s})\)
\State Forward \(\hat{x}\) into model \(Q\) and optimize \(Q\)
\EndFor
\EndProcedure
\State For every \(C\) epochs, copy parameters \(\hat{Q} \gets Q\)
\Else
\Procedure{Evaluate model \(Q\)}{}:
\State Project point cloud \(c\) into image \(x\)
\State Get mask $s = S(x)$
\State Get DWT coefficients $h$ with $x = f(h)$
\State Set masked image \(\hat{x} := f(h \odot s)\)
\State Forward $\hat{x}$ into model \(Q\) for evaluation
\EndProcedure
\EndIf
% \EndFor
\EndFor 
\end{algorithmic}
\end{algorithm*}

\end{document}